\documentclass[journal]{IEEEtran}
\usepackage{amsfonts}
\usepackage{graphicx}
\usepackage{array}
\usepackage{cite}
\usepackage{indentfirst}
\usepackage{amsmath}
\usepackage{multirow}
\usepackage{multicol}
\usepackage{float}

\usepackage{epstopdf}
\ifCLASSINFOpdf
\else
\fi
\usepackage{authblk}

\begin{document}
%
\title{3D Gaussian Splatting for Large-scale Surface Reconstruction from Aerial Images}
%
%
%

\author{Yuanzheng Wu,
        Jin Liu,
        Shunping Ji
\thanks{S. Ji, Y. Wu are with the School of Remote Sensing and Information Engineering, Wuhan University, Wuhan, HB 430079, China (e-mail: ji
 shunping@whu.edu.cn; yuanzhengwu@whu.edu.cn).}
 \thanks{J. Liu is with the School of Communication Engineering, Hangzhou Dianzi University, Hangzhou, ZJ 310018, China (e-mail: jinliu@hdu.edu.cn).}
 }

\maketitle

\begin{abstract}
Recently, 3D Gaussian Splatting (3DGS) has demonstrated excellent ability in small-scale 3D surface reconstruction. However, extending 3DGS to large-scale scenes remains a significant challenge. To address this gap, we propose a novel 3DGS-based method for large-scale surface reconstruction using aerial multi-view stereo (MVS) images, named Aerial Gaussian Splatting (AGS). First, we introduce a data chunking method tailored for large-scale aerial images, making 3DGS feasible for surface reconstruction over extensive scenes. Second, we integrate the Ray-Gaussian Intersection method into 3DGS to obtain depth and normal information. Finally, we implement multi-view geometric consistency constraints to enhance the geometric consistency across different views. Our experiments on multiple datasets demonstrate, for the first time, the 3DGS-based method can match conventional aerial MVS methods on geometric accuracy in aerial large-scale surface reconstruction, and our method also beats state-of-the-art GS-based methods both on geometry and rendering quality.
\end{abstract}

%
\IEEEpeerreviewmaketitle

\section{Introduction}
%
%
%
%
\IEEEPARstart
Large-scale surface reconstruction has long been a focal point of interest in both academic research and industrial applications, particularly in domains such as aerial surveying \cite{meyer2015optimizing} \cite{singh20143d} and smart city development \cite{danilina2018smart} \cite{buyukdemircioglu2020reconstruction} \cite{khan2020challenges} \cite{lim2019identifying}. Recently, NeRF-based methods \cite{mildenhall2021nerf}\cite{pumarola2021d}\cite{ma2022deblur} have been extensively researched for image rendering, especially when applied to small-scale foreground targets. Additionally, these methods have shown potential for surface reconstruction \cite{wang2021neus}\cite{li2023neuralangelo}. However, the high computational cost of volumetric rendering in NeRF-based approaches makes them impractical for large-scale scenes. The emergence of the 3D Gaussian Splatting (3DGS) technology \cite{kerbl20233d} offers an alternative solution. In contrast to NeRF-based methods, 3DGS uses 3D Gaussian primitives instead of the implicit radiance field learned through Multi-Layer Perceptions (MLPs) to represent a scene. The training process is achieved by optimizing parameters such as positions, rotations, and scales of these Gaussian primitives. This approach markedly reduces computational requirements and enables more efficient scene rendering and reconstruction, making it a viable solution for large-scale reconstruction using aerial images captured by airplanes or unmanned aerial vehicles (UAVs).

Although 3DGS has demonstrated impressive capabilities in high-fidelity novel view synthesis and real-time rendering \cite{yu2024mip} \cite{lu2024scaffold}, achieving large-scale aerial surface reconstruction with geometric precision comparable to or exceeding that of mainstream conventional multi-view stereo (MVS) methods \cite{schoenberger2016sfm} \cite{schoenberger2016mvs} \cite{openmvs2020} or deep learning-based approaches \cite{yao2018mvsnet} \cite{liu2020novel} remains a significant challenge. First, the extensive scenes and a large number of aerial images impose substantial computational demands, often leading to out-of-memory issues on modern GPUs. Second, it is challenging to determine the intersection between a Gaussian and a ray, making it difficult to obtain precise depth and normal vector information. Applying 3DGS directly to surface reconstruction tasks frequently results in low-precision surface models. Third, the original 3DGS \cite{kerbl20233d} relies solely on image-related loss for optimization, which skews the Gaussian distribution toward high-fidelity image rendering at the expense of surface geometry accuracy. To address these challenges, we propose a novel framework based on 3DGS for large-scale surface reconstruction from aerial images. To the best of our knowledge, this is the first application of 3DGS methods to aerial images for achieving high-precision surface reconstruction.

The proposed large-scale surface reconstruction framework, named Aerial Gaussian Splatting (AGS), builds upon 3D Gaussian Splatting (3DGS) \cite{kerbl20233d} as its baseline. While 3DGS performs well for small-scale scenes, it struggles with large-scale environments due to high memory demands. To address the memory challenges, we tackle the problem by partitioning scenes for parallel training. A key issue in aerial scene partitioning is the uneven distribution of point clouds—some regions suffer from sparse views and points, while others are oversaturated with redundant views. The proposed method overcomes this by adopting the chunking method from VastGaussian \cite{lin2024vastgaussian}, partitioning the scene based on camera positions and expanding the boundaries of each data block. Additionally, to ensure the inclusion of more suitable viewpoints within each block, we develop a viewpoint selection and culling strategy. This approach enhances scene optimization by incorporating relevant viewpoints and discarding less useful ones, resulting in more efficient and balanced processing across blocks. We refer to the entire chunking method as Adaptive Aerial Scene Partitioning.

Due to the inability to precisely determine the intersection between a Gaussian and a ray, estimating accurate depth and normal vector information for each Gaussian primitive is a significant challenge, which limits the application of effective geometric constraints. To address this, the proposed method adopts the Ray-Gaussian Intersection (RGI) approach from \cite{keselman2022approximate}\cite{yu2024gaussian}, which accurately retrieves both depth and normal vector information. Once this information is acquired, we apply depth and normal consistency constraints\cite{huang20242d} to enhance accuracy. Furthermore, to improve geometric consistency across different views, similar to MVS methods \cite{liu2023deep}\cite{xu2019multi}\cite{dong2022patchmvsnet}, we introduce a multi-view geometric consistency strategy. This strategy calculates the error of the rendered depth maps through projection and reprojection. By doing so, we improve the reconstruction of surface details, resulting in more accurate geometric alignment across different views.

We evaluate the geometric accuracy of our framework on the WHU-OMVS \cite{liu2023deep} and Tianjin aerial datasets. The experimental results show that the proposed method outperforms the existing 3DGS-based approaches and, in some cases, even surpasses the open-source MVS software Colmap \cite{schoenberger2016sfm} \cite{schoenberger2016mvs} and OpenMVS \cite{openmvs2020}. Furthermore, we validate the rendering quality of the proposed method on the WHU-OMVS, Mill-19 \cite{turki2022mega} and UrbanScene3D \cite{lin2022capturing} datasets, where our method achieves better performance than other GS-based methods. 

Our contributions are summarized as follows:

\begin{itemize}
  \item We introduce a novel large-scale aerial surface reconstruction and rendering method based on 3DGS. 
  \item We adapt 3DGS for large-scale scenes by employing a chunking method based on the VastGaussian and designing a viewpoint selection and culling strategy to optimize the chunking process.
  \item We introduce the ray-gaussian intersection strategy and multi-view geometric consistency constraints into the framework, which significantly improves geometric accuracy.
  \item We conduct experiments on multiple datasets, and the results demonstrate that the proposed method achieves high-quality surface reconstruction and delivers high-fidelity rendering.
\end{itemize}
\section{Related work}
\subsection{Novel View Synthesis}
Recent advances in Neural Radiance Fields (NeRF) \cite{mildenhall2021nerf} have significantly influenced the Novel View Synthesis (NVS) domain by employing neural networks to learn and render high-quality 3D representations of continuous volumetric scenes from images taken from multiple viewpoints. NeRF achieves high-fidelity scene representation by predicting density and color. However, the huge computational demands and the time required for both training and rendering present significant challenges for real-time rendering in large-scale scenarios. To address NeRF's computational demands, Plenoxels \cite{fridovich2022plenoxels} uses 3D sparse grids to represent scene points, reducing computational complexity and storage requirements while enhancing speed compared to vanilla NeRF. However, the use of voxel grids in Plenoxels leads to a degradation of fine details. Efforts like Mip-NeRF \cite{barron2021mip} and Tri-MipRF \cite{hu2023tri} enhance rendering quality with multi-scale representations and anti-aliasing, achieving better visual quality without sacrificing efficiency.

More recently, 3DGS \cite{kerbl20233d} has gained attention for its ability to achieve high-fidelity and real-time rendering. Subsequent work, such as Mip-Splatting \cite{yu2024mip}, employs 3D smoothing filters to improve rendering quality. Scaffold-GS \cite{lu2024scaffold} leverages anchor points to regulate local 3D Gaussian distributions, allowing for real-time adjustments to both the distribution and density of Gaussians.

\begin{figure*}
    \centering
    \includegraphics[width=0.9\linewidth]{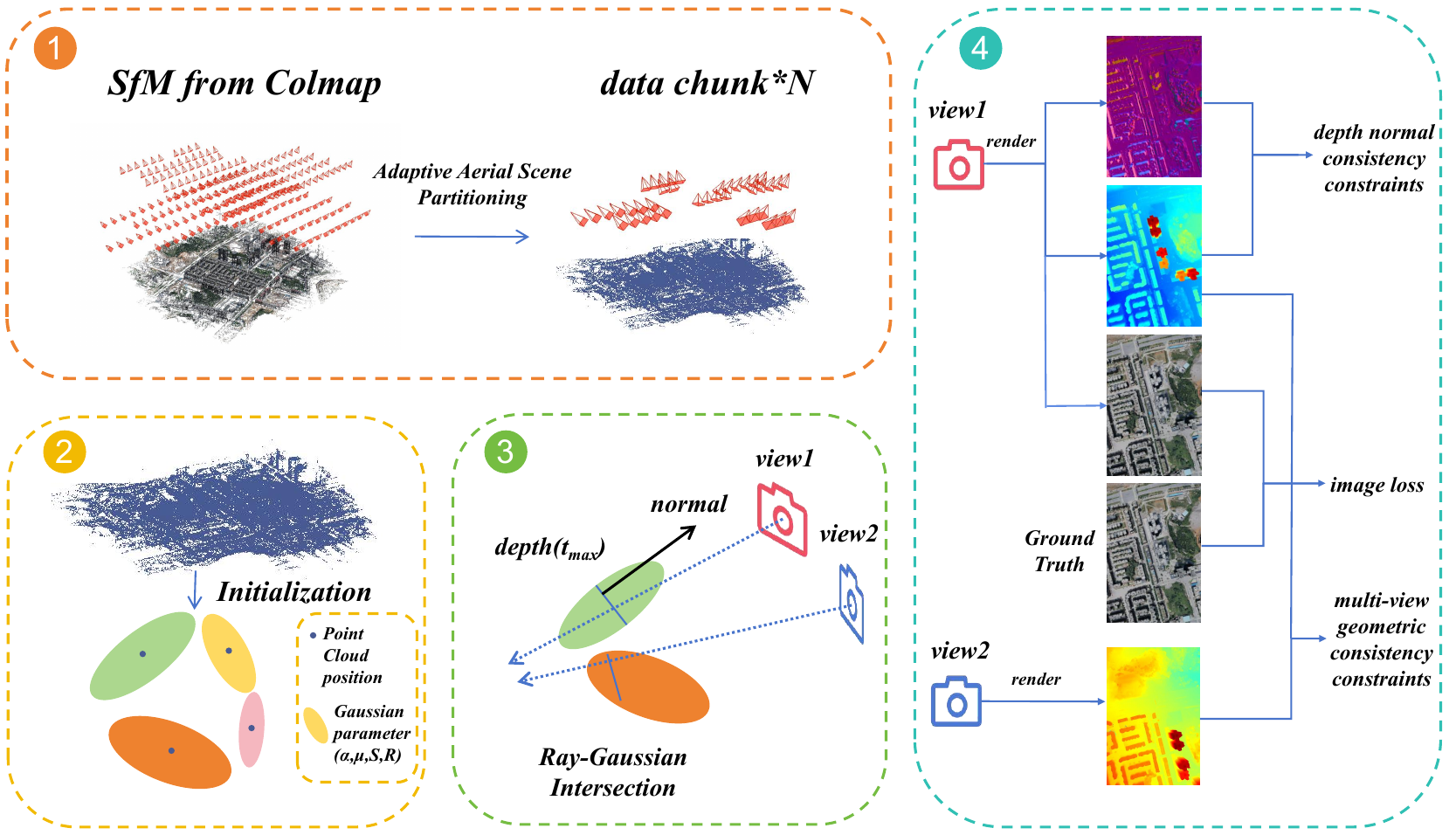}
    \caption{The overview of AGS Framework. (1) The SfM sparse point clouds and views are divided into N data blocks. (2) The point clouds in each block are used to initialize the 3D Gaussians. (3) The ray-gaussian intersection technique is applied to obtain depth and normal vector information. (4) The depth map and normal map are utilized to compute the depth normal consistency constraints and multi-view geometric consistency constraints.}
    \label{fig:method}
\end{figure*}
\subsection{Surface Reconstruction}

Image-based reconstruction has advanced considerably over the decades. Semi-global matching (SGM) \cite{hirschmuller2007stereo} and patch-based methods \cite{bleyer2011patchmatch} have been widely used for dense image matching. Complete surface reconstruction solutions, such as Colmap \cite{schoenberger2016sfm}\cite{schoenberger2016mvs} and OpenMVS \cite{openmvs2020}, utilize dense matching techniques to generate dense point clouds or triangulated meshes. With the rapid development of deep learning, learning-based MVS methods \cite{yao2018mvsnet} \cite{liu2020novel} have been developed to predict depth maps from multi-view images, and some of these methods are integrated into comprehensive surface reconstruction frameworks \cite{liu2023deep}.

More recently, several works have attempted to apply NeRF or 3DGS-based methods to surface reconstruction, mainly for small-scale or foreground objects. NeuS \cite{wang2021neus} combines the strengths of both volumetric and surface rendering by optimizing Signed Distance Functions (SDF) and a color field to reconstruct fine surface details, but it requires substantial computational resources and inference time. Neuralangelo \cite{li2023neuralangelo} improves surface reconstruction fidelity by combining multi-resolution 3D hash grids with neural surface rendering, but it also demands significant computational power.

The emergence of 3DGS \cite{kerbl20233d} has introduced new approaches for surface reconstruction. SuGaR \cite{guedon2024sugar} compresses 3D Gaussian spheres into approximate 2D ellipses during training and utilizes Poisson reconstruction to extract continuous mesh from 3D point clouds, which are sampled based on the Gaussian density field. However, the absence of geometric constraints leads to dispersed point clouds and numerous holes in the meshes. 2D Gaussian Splatting \cite{huang20242d} replaces 3D Gaussian ellipsoids with 2D disks, addressing the limitations of vanilla 3DGS in surface reconstruction. Nonetheless, 2DGS still struggles to capture the intricate details of large-scale scenes.

At present, most GS-based surface reconstruction methods predominantly focus on small-scale foreground targets, and large-scale surface reconstruction remains largely underexplored. 
\subsection{Large Scene Reconstruction}
In rendering research, the applications of the radiance field technology have extended from rendering close-range foreground objects to large-scale scenes. Block-NeRF \cite{tancik2022block} and Mega-NeRF \cite{turki2022mega} use data chunking strategies to handle large-scale scenes, while UE4-NeRF \cite{gu2024ue4} integrates NeRF with Unreal Engine 4 for scalable rendering and scene editing. Switch-NeRF \cite{zhenxing2022switch} uses a mixture of expert models for scene decomposition, enhancing large-scale scene reconstruction.

A few studies use 3DGS for large-scale scene rendering. CityGaussian \cite{liu2024citygaussian} explores large-scale scene rendering using chunking and Level of Detail (LoD) methods to address challenges related to rendering efficiency and scalability. Similarly, VastGaussian \cite{lin2024vastgaussian} addresses lighting effects on rendering, further enhancing the visual quality of large-scale scenes. 

Most of the aforementioned NeRF and 3DGS-based studies primarily focus on image rendering, with few addressing large-scale scene surface reconstruction. In contrast, our work seeks to extend 3DGS-based methods to surface reconstruction from large-scale aerial MVS images. We borrow the idea of block chunking in the large-scale rendering method VastGaussian \cite{lin2024vastgaussian} and develop a viewpoint selection and culling strategy to bridge the huge computational resource demand and limited GPU capacity. Additionally, we introduce the ray-gaussian intersection method \cite{keselman2022approximate}\cite{yu2024gaussian} to determine the intersection point between the Gaussian and the ray, thereby enabling the acquisition of accurate depth and normal vector information. Furthermore, we apply multi-view geometric consistency constraints to ensure geometric consistency across different views, enhancing high-fidelity and high-precision large-scale scene reconstruction.

\section{Preliminaries}
3DGS\cite{kerbl20233d} represents scenes explicitly using a large number of 3D Gaussian primitives (ellipsoids). Each Gaussian primitive is presented by four types of parameters that require optimization: position, covariance, opacity, and spherical harmonics (SH) coefficients. Using these four parameters, the $\alpha$ -blending algorithm is employed to render a new image from these 3D Gaussians. Specifically, for a pixel $p_i$ in the rendered image, the color of $p_i$ can be obtained by: 
\begin{equation}
C(p_i)=\sum_{i\in N} \alpha_i c_i \prod_{j=1}^{i-1} ( 1 - \alpha_j )
\end{equation}

Where $N$ represents the number of all Gaussians, $c_i$ is the view-dependent color of the $i$-th Gaussian, derived from spherical harmonics coefficients, $\alpha_i$ is determined by the Gaussian distribution, and the Gaussian's opacity $\sigma_i$, as seen in Equation (2).
\begin{equation}
\alpha_i = \sigma_i exp(- \frac{1}{2} (p - \mu_i) ^ T\Sigma_{i}^{-1}(p-\mu_i))
\end{equation}

The parameters $\Sigma$ is given by $\Sigma=RSS^TR^T$, where $R\in \mathbb{R}^{3\times3}$ is the rotation matrix and $S\in\mathbb{S}^{3\times1}$ is the scale matrix.

When calculating the depth map, we accumulate the blending weights $\alpha$ at pixel $p$. Once the cumulative value exceeds 0.5, the depth of the current Gaussian is assigned as the depth value for that pixel. In this paper, the depth of a Gaussian is calculated by the ray-gaussian intersection method\cite{keselman2022approximate}\cite{yu2024gaussian}.

\begin{figure*}[h]
    \centering
    \includegraphics[width=0.9\linewidth]{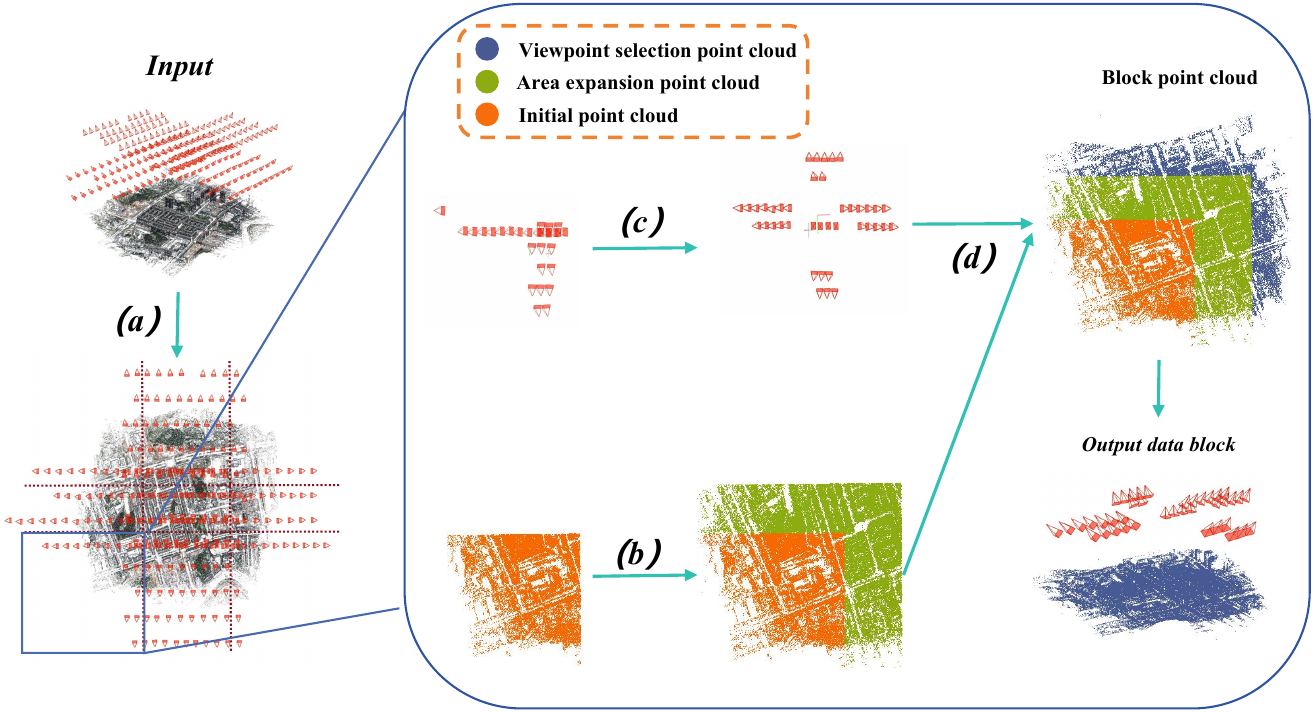}
    \caption{Overview of Adaptive Aerial Scene Partitioning strategy. (a) The entire scene is divided into N regions based on camera positions. (b) The boundaries of each region are expanded. (c) Viewpoints (i.e., cameras) are selected and culled. (d) All points visible from the selected viewpoints within each data block are added to the block’s point cloud.} 
    \label{fig:chunking}
\end{figure*}
\section{Method}
In this section, we present Aerial Gaussian Splatting (AGS), a surface reconstruction framework based on 3DGS, specifically designed for aerial MVS images. Compared to the original 3DGS \cite{kerbl20233d}, the proposed method incorporates several innovative and necessary modules. First, we introduce adaptive aerial scene partitioning to divide large-scale scenes effectively and ensure optimal merging at the final step. Furthermore, the ray-gaussian intersection technique\cite{keselman2022approximate}\cite{yu2024gaussian} is employed to obtain accurate depth and normal vector information. Finally, multi-view geometric consistency constraints are incorporated to improve the reconstruction quality. The framework is shown in Fig. \ref{fig:method}.

\subsection{Adaptive Aerial Scene Partitioning}
In the field of aerial photogrammetry, certain basic principles guide the partitioning of a scene into blocks \cite{liu2023deep} to alleviate the computational burden. In this paper, our method is inspired by VastGaussian \cite{lin2024vastgaussian} and involves three main steps. The first two steps are derived from VastGaussian. The process, as shown in Fig. \ref{fig:chunking}, begins by dividing the scene based on camera positions and extending the block boundaries. Specifically, the scene, containing a total of n viewpoints, is divided into M×N blocks. First, the viewpoints are horizontally divided into M blocks, with each block containing n/M viewpoints. Then, these M blocks are further divided vertically, resulting in each block containing n/(M×N) viewpoints. After this initial partitioning, the point clouds within each region are aggregated into distinct point cloud blocks. To minimize artifacts across the scene, each region is extended by a certain proportion as in VastGaussian, ensuring that each block is adequately optimized.

However, a coarse data block partitioning method based solely on camera positions may result in insufficient optimization within each block due to suboptimal viewpoints. To address this, we develop a viewpoint selection and culling strategy, as shown in Fig. \ref{fig:chunking}(c) and (d). This strategy removes erroneous viewpoints from the data block and supplements it with additional effective viewpoints. This process involves projecting all point clouds within a data chunk onto all images and calculating projection scores to determine the suitability of each viewpoint. If a point falls within the central scope (here 70\%) of an image, the projection score for that image is incremented by one. For each region, the top N images with the highest scores are selected as viewpoints to optimize the data block. Finally, to ensure sufficient points for initialization and to mitigate artifacts, the sparse point cloud generated from the SfM of all new viewpoints is incorporated into the data block.
\begin{figure}[h]
    \centering
    \includegraphics[width=0.9\linewidth]{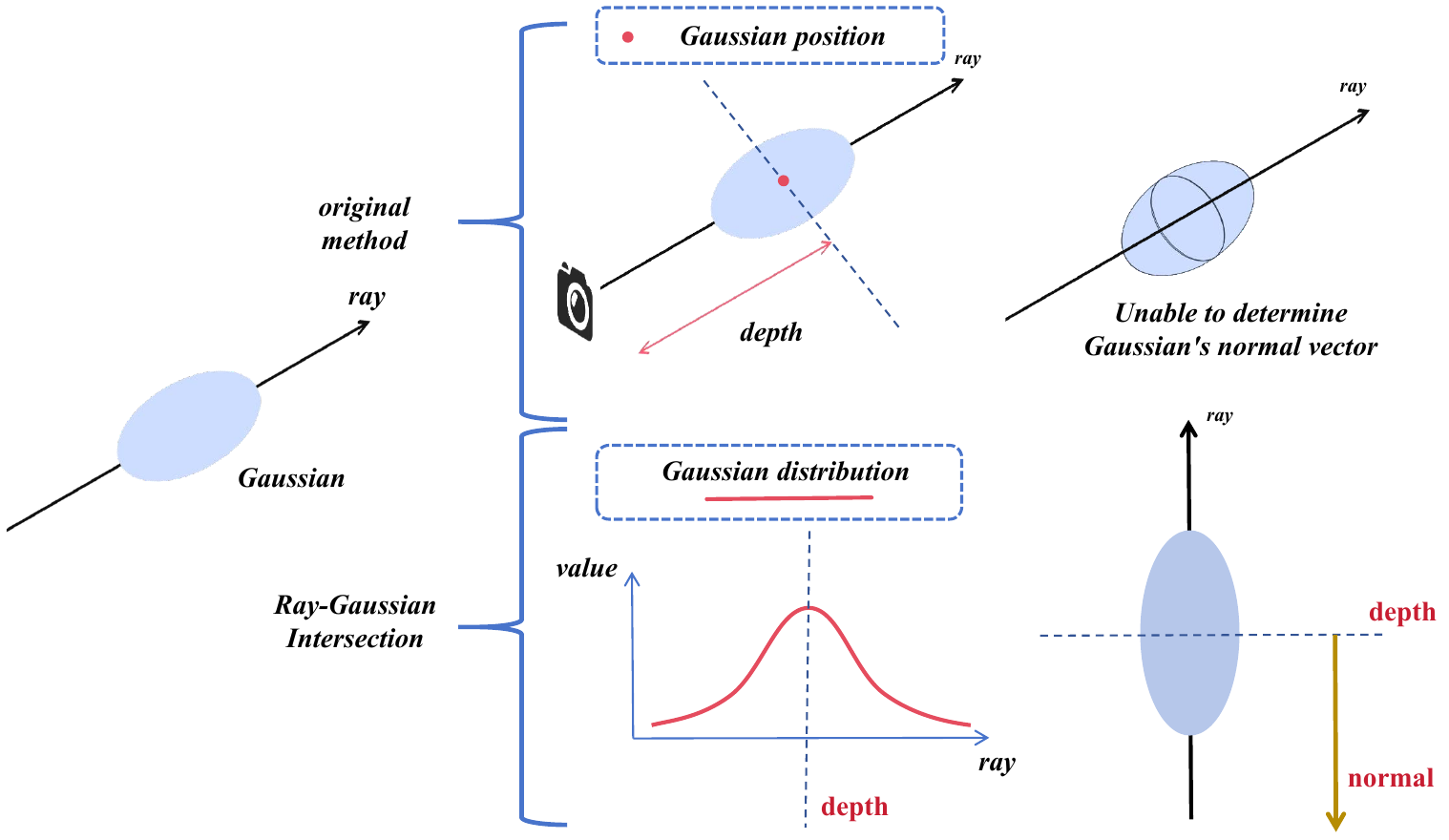}
    \caption{Ray-Gaussian Intersection. By calculating the maximum Gaussian value along the ray, we can obtain accurate depth and normal vector information for the Gaussian.}
    \label{fig:RGI}
\end{figure}

\subsection{Ray-Gaussian Intersection}
In vanilla 3DGS, the depth of each Gaussian primitive is assigned based on its distance from the screen, and accurate normal vector information is not provided. However, for surface reconstruction tasks, reliable depth information and normal vector data are essential for geometric constraints. To address these issues, we introduce the ray-gaussian intersection technique \cite{keselman2022approximate} \cite{yu2024gaussian}, as shown in Fig. \ref{fig:RGI}, to improve the accuracy of surface reconstruction. The intersection point $t_{max}$ between a Gaussian primitive and a ray, corresponding to the maximum Gaussian value along the ray, can be computed as follows \cite{yu2024gaussian}:

\begin{equation}
    t_{max}=-\frac{o_g^Tr_g}{r_g^Tr_g}
\end{equation}

$o_g$, $r_g$ are $o$ (the camera center) and $r$ (the ray direction) converted into the Gaussian local coordinate system. An arbitrary point along the ray is defined as $x = o + tr$, where $t$ is the depth of the ray.

Once the depth value is computed, the Gaussian's normal is derived as the normal of the intersection plane relative to the given ray direction. For image rendering, rather than projecting 3D Gaussians onto 2D screen space as done in the original 3DGS, we utilize the ray-gaussian intersection method to determine the intersection point between the Gaussian and the ray, allowing us to compute the contribution of a Gaussian to a given ray in 3D space.

After obtaining the depth and normal vector information, we apply 2DGS's depth normal consistency constraints\cite{huang20242d}. Specifically, this involves calculating the error between the normal map and the gradient values derived from the depth map, which is used as the loss function.

\begin{equation}
    L_n = \sum_i \omega_i (1-n_i^TN)
\end{equation}

Here, $i$ represents the Gaussian index, $\omega_i$ denotes the blending weight, $n_i$ represents the normal vector of the Gaussian, and $N$ is the normal vector calculated from the depth map. The normal vector in the depth map at a given point is computed as Eq.(\ref{eq:tidu}), in which $p$ represents the pixel coordinates and $\nabla$ represents the gradient calculation. 

\begin{equation}
    N(x,y)=\frac{\nabla_x p \times \nabla_y p}{|\nabla_x p \times \nabla_y p|}
    \label{eq:tidu}
\end{equation}
\subsection{Multi-view geometric consistency constraints}
\begin{figure}[t]
    \centering
    \includegraphics[width=0.8\linewidth]{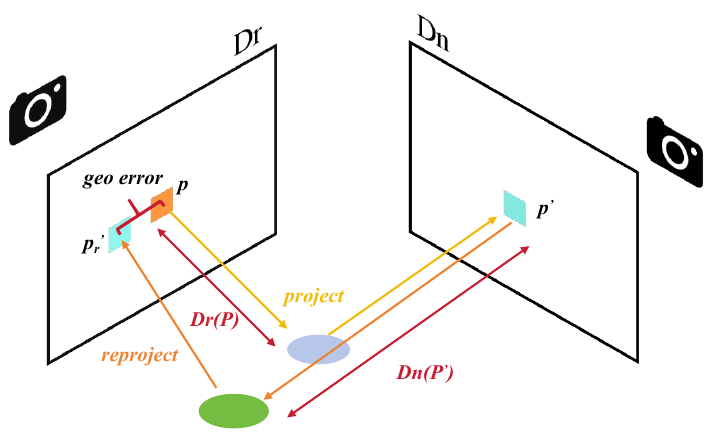}
    \caption{Multi-view geometric consistency constraints. The multi-view geometric consistency constraints are modeled as the error between projection and reprojection of depth map across multiple views.}
    \label{fig:mulgeo}
\end{figure}
As 3DGS-based surface construction is still in its early stages, certain beneficial empirical approaches, such as multi-view geometric consistency constraints, have yet to be fully developed. In this study, we introduce multi-view geometric consistency constraints to ensure geometric coherence across multiple views. As shown in Fig. \ref{fig:mulgeo}, we render the depth maps for two adjacent viewpoints $V_r$ and $V_n$, denoted as $D_r$ and $D_n$. Firstly, a pixel $P$ in the reference view $V_r$ is projected onto the adjacent view $V_n$ through its depth value $Dr(P)$ and the intrinsic and extrinsic parameters, yielding the projected point $P^\prime$ in $V_n$. Subsequently, the projected point $P^\prime$ is reprojected onto the reference view based on its rendered depth value $Dn(P^\prime)$, resulting in the reprojected pixel $P_r^\prime$. The distance between the coordinates of $P$ and $P_r^\prime$ is calculated as the geometric consistency constraints: 
\begin{equation}
    L_{geo}=\frac{1}{V} \sum_{P \in V }\| P-P_r^\prime \|
    \label{eq:mulgeo}
\end{equation}

When calculating the loss, only the non-zero values are averaged, as shown in Eq.(\ref{eq:mulgeo}), where $V$ represents the valid pixels. To reduce the impact of occlusion, a distance threshold T (It is usually set to 1.) is applied to identify valid pixels, with distances $\|P-P_r^\prime\|$ exceeding T set to zero.

\begin{figure*}[htb]
    \centering
    \includegraphics[width=0.8\linewidth]{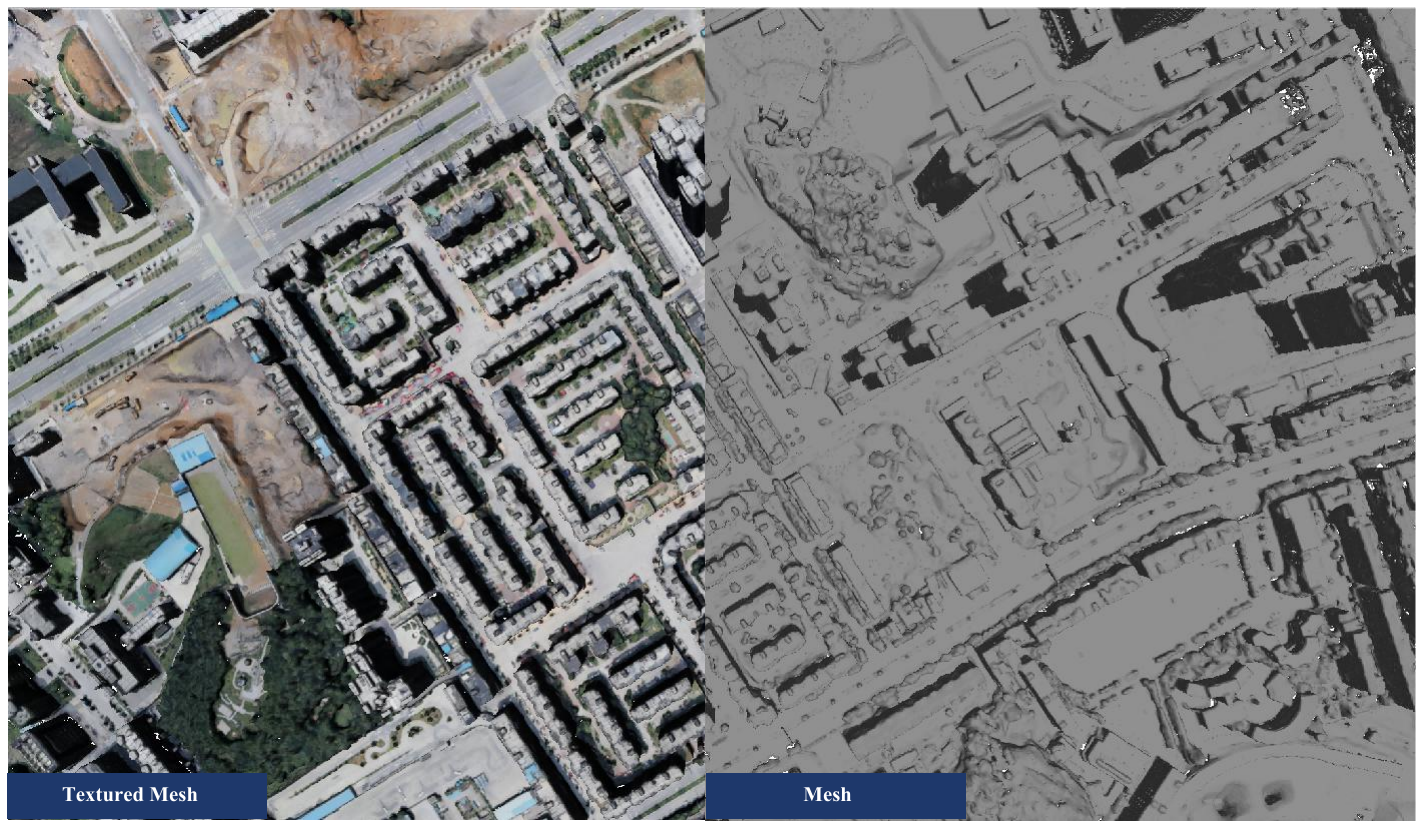}
    \caption{Surface reconstruction results of WHU-OMVS dataset.}
    \label{fig:omvs_mesh}
\end{figure*}
\subsection{Merging}
After parameters in each data block are optimized separately, all blocks are merged to form a coherent scene. This is achieved by removing the expanded regions of each block prior to merging.

\begin{table*}[htbp]
    \centering
    \caption{The quantitative results of surface reconstruction on the WHU-OMVS dataset. The best results are highlighted in \textbf{bold}, and the second-best results are underlined.}
    \begin{tabular}{>{\centering\arraybackslash}p{3cm} | >{\centering\arraybackslash}p{2cm} >{\centering\arraybackslash}p{2cm} >{\centering\arraybackslash}p{2cm} >{\centering\arraybackslash}p{2cm} >{\centering\arraybackslash}p{2cm}}
        \hline
        method & PAG$_{0.6m}$(\%) & PAG$_{0.8m}$(\%) & PAG$_{1.0m}$(\%) & MAE(m) & RMSE(m)\\ 
        \hline
         OpenMVS & 71.91 & 77.81 & 81.18 & 0.520 & \underline{1.041}\\ 
         Colmap & 71.32 & 77.44 & 81.36 & 0.623 & 1.287\\  
         GOF & \underline{79.80} & \underline{83.83} & \underline{86.03} & \underline{0.476} & 1.155\\
         3DGS & 65.01 & 73.09 & 74.98 & 0.749 & 1.366\\
         2DGS & 56.88 & 65.57 & 69.42 & 0.906 & 1.483\\
         proposed & \textbf{82.58} & \textbf{87.07} & \textbf{89.50} & \textbf{0.451} & \textbf{1.012}\\
         \hline
    \end{tabular}
\label{tb:OMVS_surface}
\end{table*}
\section{Experiment}
\subsection{Dataset}
\textbf{WHU-OMVS: }This dataset covers an area of Guizhou, China, with a ground resolution of 10 cm. The images are captured using a camera rig with one nadir and four oblique viewpoints, totaling 268 images, each with a resolution of 3712×5504 pixels. The flight height is 550 meters, covering an area of 850×700 $m^2$. Due to GPU memory limitations, we apply a 4x downsampling to the images during the training of all methods, as done in previous studies \cite{liu2024citygaussian} \cite{lin2024vastgaussian}\cite{zhenxing2022switch}\cite{turki2022mega} for rendering and reconstruction. The depth map is used as ground truth, and we evaluate the geometric accuracy of the proposed method based on the rendered depth map.

\textbf{Tianjin Dataset: }This dataset is captured by a camera rig with one nadir and four oblique viewpoints over Tianjin city, China, with an image size of 3840×2560 pixels and a ground resolution of 20 cm captured at a height of 200 meters. The dataset consists of 342 images and covers an area of 400×350 $m^2$. The image overlap is 80\% along the heading direction and 60\% in the side direction. The ground truth data is derived from LiDAR point cloud scans. We apply a 4x downsampling operation to all images.

\textbf{Mill-19 and UrbanScene3D:} We apply the proposed method to three open-source large-scale scenes: Rubble and Building from the Mill-19 dataset \cite{turki2022mega} and Residence from the UrbanScene3D dataset \cite{lin2022capturing}, containing 1,678, 1,940, and 2,582 images, respectively. Following previous rendering methods \cite{liu2024citygaussian} \cite{lin2024vastgaussian}\cite{zhenxing2022switch}\cite{turki2022mega}, we perform a 4x downsampling operation to the input image during training. Due to the absence of ground truth, we conduct a qualitative analysis of the surface reconstruction results from these datasets.
\subsection{Implementation}
When training the proposed method, we first perform Manhattan alignment on the target scene, aligning the y-axis to be perpendicular to the ground to facilitate the chunking process. Each block is expanded by 20\%. During training, each data block is independently optimized for 50,000 iterations. The densification process begins after 500 iterations and ends at 30,000 iterations. The multi-view geometric consistency constraints and depth normal consistency constraints are introduced at 7,000 iterations. Experiments are performed on the RTX4090. Other settings remain consistent with those used in the original 3DGS method \cite{kerbl20233d}. For the sparse point cloud generation, we use the SfM module in Colmap \cite{schoenberger2016mvs} \cite{schoenberger2016sfm}. For surface reconstruction, we follow the 2D Gaussian Splatting approach\cite{huang20242d} and utilize the Truncated Signed Distance Function (TSDF) \cite{werner2014truncated}. When evaluating 3DGS \cite{kerbl20233d}, we set the densification interval to 250 instead of the original 100 to avoid the out-of-memory problem.
\begin{figure*}[h]
    \centering
    \includegraphics[width=1\linewidth]{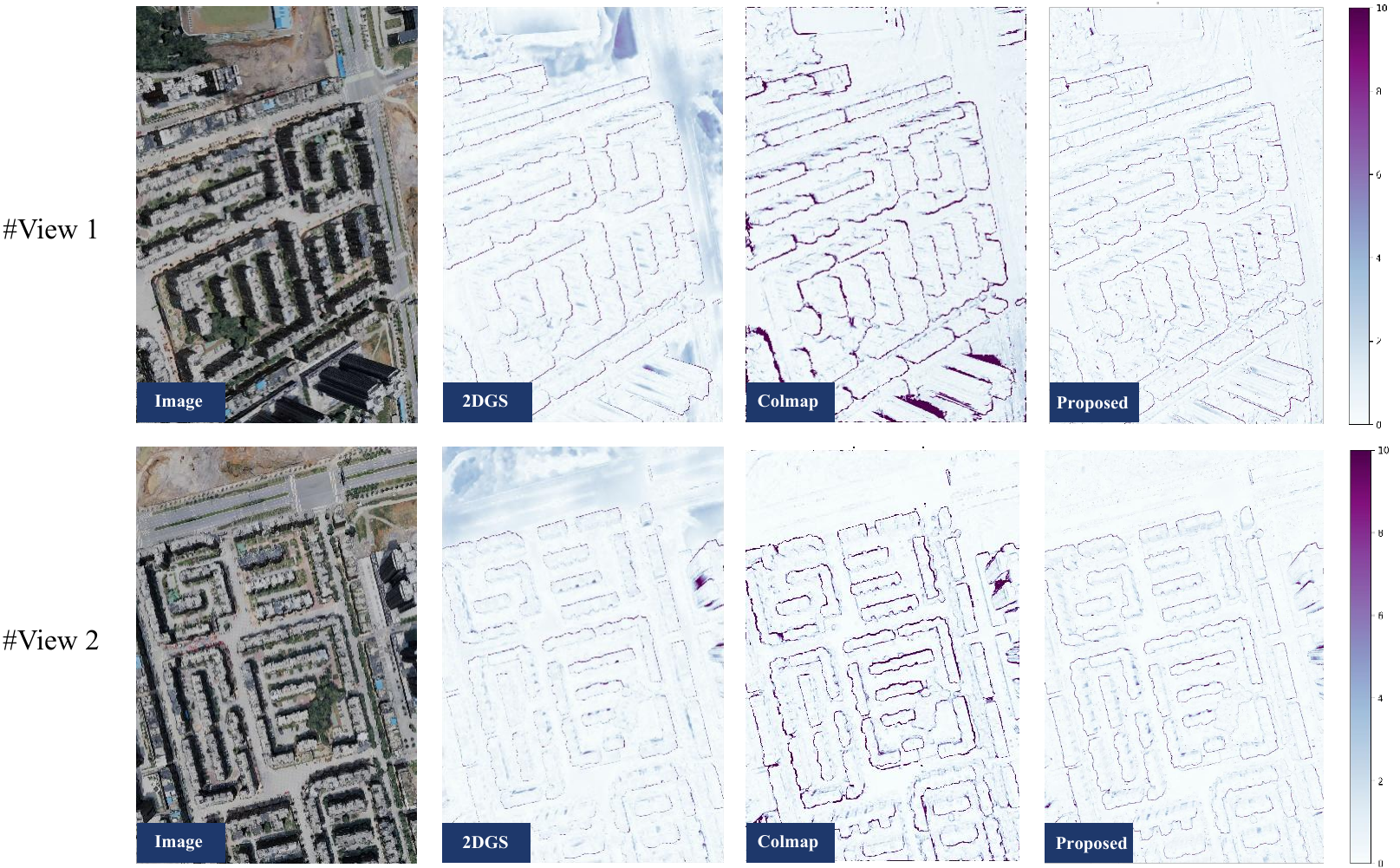}
    \caption{Qualitative comparisons with different methods. The image on the left represents the original image from the viewpoint, while the three images on the right depict the error maps between the predicted depth values and the ground truth. The error maps' bands indicate the errors' magnitude, with darker colors representing larger errors.}
    \label{fig:OMVS_error}
\end{figure*}



\subsection{Results}
We conduct surface reconstruction experiments on the WHU-OMVS \cite{liu2023deep}, Tianjin, Mill-19\cite{turki2022mega} and UrbanScene3D\cite{lin2022capturing} dataset. Due to the limited number of 3DGS-based surface reconstruction methods capable of handling scenes with significant depth variation, we select three for comparison: 2D Gaussian Splatting \cite{huang20242d}, 3D Gaussian Splatting \cite{kerbl20233d}, and Gaussian Opacity Fields \cite{yu2024gaussian}. Additionally, we include comparisons with the widely recognized open-source MVS software Colmap \cite{schoenberger2016mvs}\cite{schoenberger2016sfm} and OpenMVS \cite{openmvs2020}.

Beyond surface reconstruction, we also evaluate the rendering quality of the proposed method in Mill-19, UrbanScene3D, and WHU-OMVS. For rendering comparisons, we select four state-of-the-art methods for large-scale rendering: Mega-NeRF \cite{turki2022mega}, Switch-NeRF \cite{zhenxing2022switch}, GP-NeRF \cite{li2024gp}, and CityGaussian \cite{liu2024citygaussian}.

\subsubsection{Surface reconstruction}\quad

\begin{table*}
    \centering
    \caption{The quantitative results of surface reconstruction on the Tianjin dataset. The best results are highlighted in \textbf{bold}, and the second-best results are underlined.}
    \begin{tabular}{>{\centering\arraybackslash}p{2cm} | >{\centering\arraybackslash}p{1cm} >{\centering\arraybackslash}p{1cm} >{\centering\arraybackslash}p{1cm}  >{\centering\arraybackslash}p{1cm} >{\centering\arraybackslash}p{1cm} >{\centering\arraybackslash}p{1cm}  >{\centering\arraybackslash}p{1cm} >{\centering\arraybackslash}p{1cm} >{\centering\arraybackslash}p{1cm}}
        \hline
        \quad &\multicolumn{3}{c}{Percentage (0.6m)↑} & \multicolumn{3}{c}{Percentage (0.8 m)↑} & \multicolumn{3}{c}{Percentage (1.0 m)↑}\\
        \hline
        method & Acc. & Comp. & f-score & Acc. & Comp. & f-score & Acc. & Comp. & f-score\\ 
        \hline
        
        Colmap&\textbf{79.51}&\underline{88.79}&\textbf{83.90}&\textbf{85.55}&\underline{90.92}&\textbf{88.15}&\textbf{89.22}&\underline{92.30}&\textbf{90.73}\\
        GOF&77.05&86.53&81.51&83.09&89.06&85.97&86.73&90.90&88.76\\
        3DGS&51.27	&\textbf{93.19}&66.15&62.79&\textbf{95.49}&75.77&71.13&\textbf{96.86}&82.02\\
        2DGS&69.54&81.60&75.09&79.85&85.07&82.38&86.03&86.89&86.46\\
        proposed &\underline{79.37}	&85.92&\underline{82.52}&\underline{85.33}&89.31&\underline{86.58}&\underline{88.80}&89.31&\underline{89.06}\\
        \hline
    \end{tabular}
    \label{tb:tianjin}
\end{table*}

\textbf{Results on WHU-OMVS:}
Following the work \cite{liu2023deep}, we use the MAE, RMSE, and PAG metrics to evaluate the rendered depth map. The specific explanations of these metrics are as follows:

Mean Absolute Error (MAE): MAE measures the absolute difference between the predicted values and the ground truth. , which is calculated by:
\begin{equation}
    MAE=\frac{1}{m} \sum_{i=1}^m|y_i-\hat{y}_i|
\end{equation}
where $y_i$ represents the ground truth, $\hat{y}_i$ represents the estimated value and $m$ denotes the number of valid values. In our experiments, differences larger than 10 meters are considered invalid and excluded from the calculation.

Root Mean Square Error (RMSE): RMSE calculates the standard deviation of the differences between the estimated values and the  ground truth:
\begin{equation}
    RMSE=\sqrt{\frac{1}{m} \sum_{i=1}^m (y_i-\hat{y}_i)}
\end{equation}

Similarly, any errors exceeding 10 meters are treated as invalid and excluded from the computation.

Percentage of Accurate Grids (PAG): PAG measures the proportion of grids with absolute difference below a given threshold 
$\alpha$ relative to the total number of grids. The evaluation is conducted using three different thresholds: 0.6m, 0.8m, and 1.0m.
\begin{equation}
    PAG_\alpha = (\frac{m_\alpha}{m}\cdot100\%)
\end{equation}
$m_\alpha$ represents the valid grid, and $m$ represents the number of all grids.

The surface reconstruction results on the WHU-OMVS dataset are presented in Table \ref{tb:OMVS_surface}. The experimental results demonstrate that the proposed method achieves the best reconstruction results at PAG, MAE, and RMSE metrics. For the strictest metric $PAG_{0.6m}$, the proposed method surpass the second-best GOF by 2.78\% and the open-source software Colmap by 11.26\%, highlighting its superiority in fine-grained reconstruction. In the $PAG_{0.8m}$ and $PAG_{1.0m}$ metrics, the proposed method outperforms other approaches by large margins, confirming that its overall reconstruction quality is substantially higher than that of the competing methods. The mesh of reconstruction results are shown in Fig. \ref{fig:omvs_mesh}. The depth error distribution of different methods, shown in Fig. \ref{fig:OMVS_error}, further supports this conclusion. For example, the 2DGS results in some severe reconstruction errors in certain areas, while Colmap produces large errors along object edges and noticeable holes in some areas. In contrast, the proposed method exhibits smaller overall errors and performs exceptionally well in detailed areas.

\begin{figure}[t]
    \centering
    \includegraphics[width=1\linewidth]{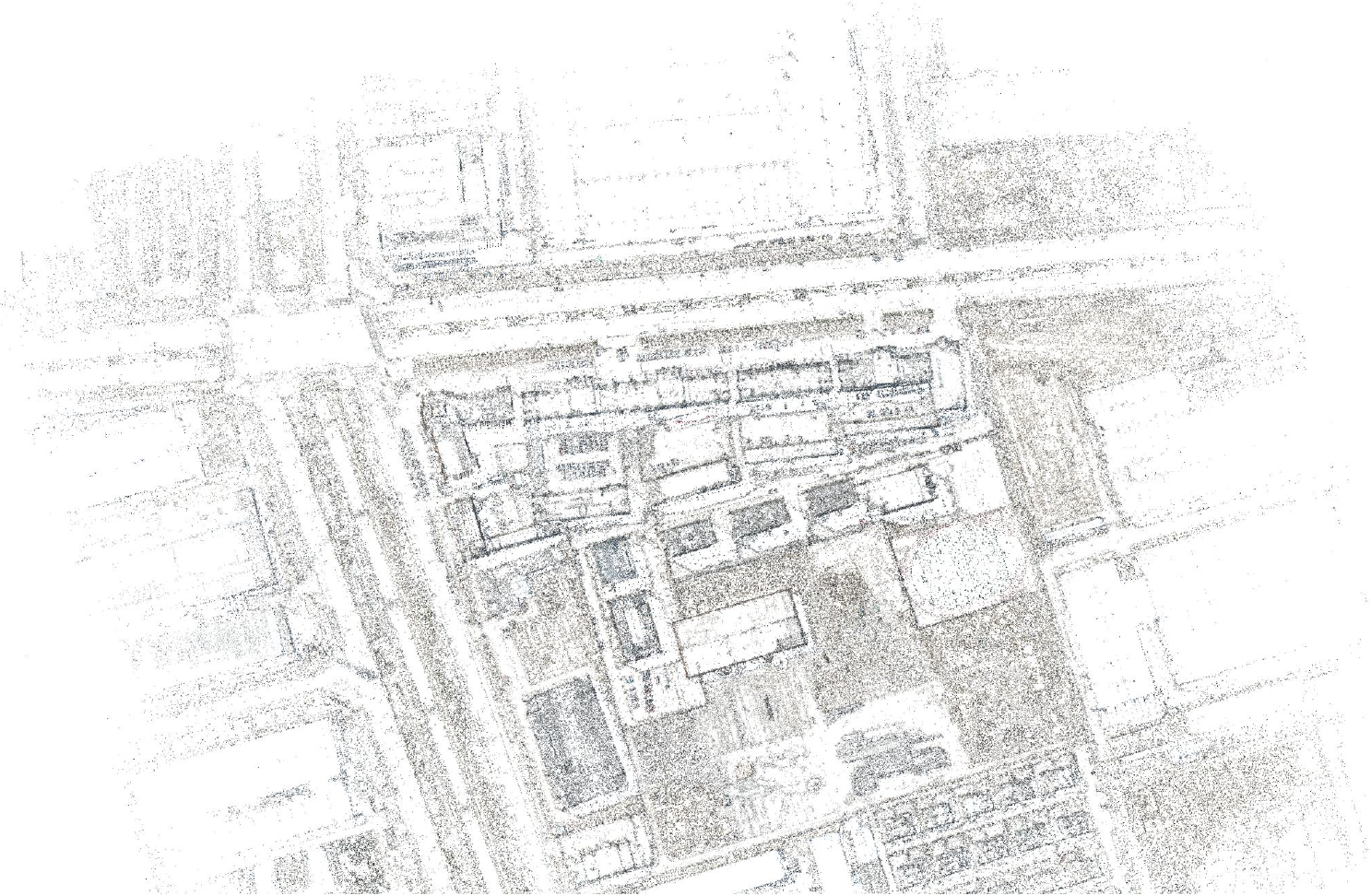}
    \caption{Sparse point cloud of the scene.}
    \label{fig:sparse_point}
\end{figure}

\textbf{Experimental Results on Tianjin:}

As the Tianjin Dataset provides only ground truth point clouds instead of pixel-wise depth maps, the accuracy of the point cloud is evaluated using the following metrics. Percentage metrics \cite{Knapitsch_Park_Zhou_Koltun_2017} are employed, with thresholds set at 0.6m, 0.8m, and 1.0m. "Accuracy" represents the distance from the reconstructed point cloud to the ground truth, while "Completeness" represents the distance from the ground truth to the reconstructed point cloud. The F-score for percentage metric is defined as the harmonic mean of accuracy and completeness.

A notable characteristic of the Tianjin dataset is the prevalence of numerous weakly textured areas, which presents significant challenges for surface reconstruction. We use SfM in Colmap to generate initial point clouds. As shown in Fig. \ref{fig:sparse_point}, certain areas—such as roads, buildings, and rooftops—exhibit notably sparse point distributions. This sparsity poses significant difficulties for 3DGS-based methods, which rely on well-distributed sparse points as the initialization for Gaussian primitives.

\begin{figure}[t]
    \centering
    \includegraphics[width=1\linewidth]{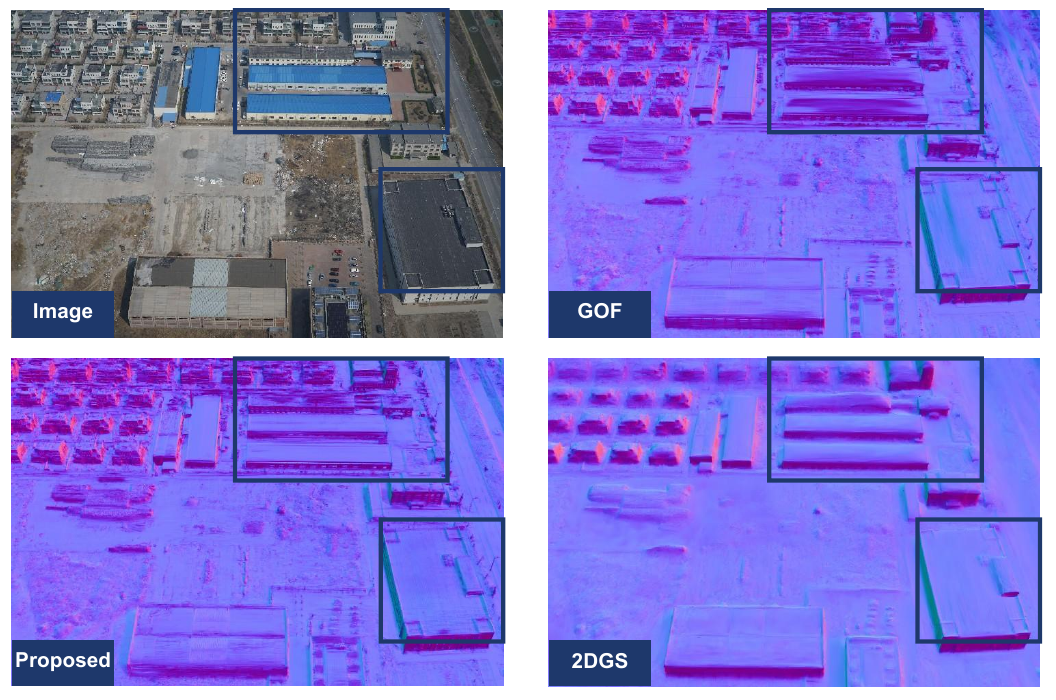}
    \caption{Qualitative comparison of normal map with 2DGS, GOF, and the proposed method.}
    \label{fig:tianjin_error}
\end{figure}

\begin{figure}[t]
    \centering
    \includegraphics[width=1\linewidth]{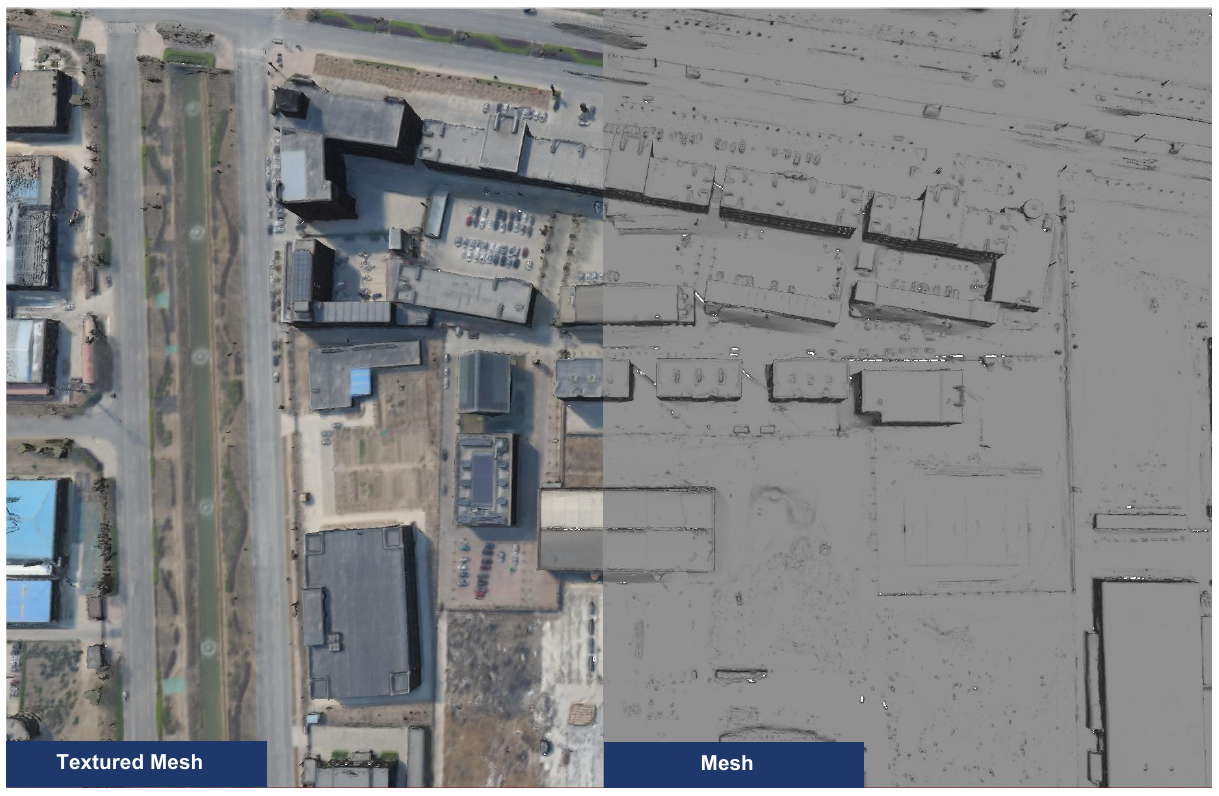}
    \caption{Surface reconstruction results of Tianjin dataset.}
    \label{fig:tianjin_mesh}
\end{figure}

As shown in Table \ref{tb:tianjin}, despite the significant challenges posed by this dataset to 3DGS-based methods, our approach achieves geometric accuracy comparable to that of Colmap, with very close metric values. Moreover, at the most strict resolution, our method outperforms other 3DGS-based methods, achieving 9.83\% higher accuracy than 2DGS, 26.10\% higher accuracy than 3DGS, and surpassing GOF by 2.32\%. Due to the point cloud count after fusion far exceeding that of the ground truth, the completeness metrics do not effectively reflect the true quality of the reconstruction. As seen in the normal map shown in Fig. \ref{fig:tianjin_error}, the proposed method exhibits a much cleaner and smoother result than other methods. In the marked regions, GOF shows significant noise, and 2DGS lacks fine details, whereas the proposed method maintains a clean and smooth appearance, showcasing its superiority in handling weak textures and complex geometries. The mesh of reconstruction results is shown in Fig. \ref{fig:tianjin_mesh}.

\begin{figure*}[htb]
    \centering
    \includegraphics[width=1.0\linewidth]{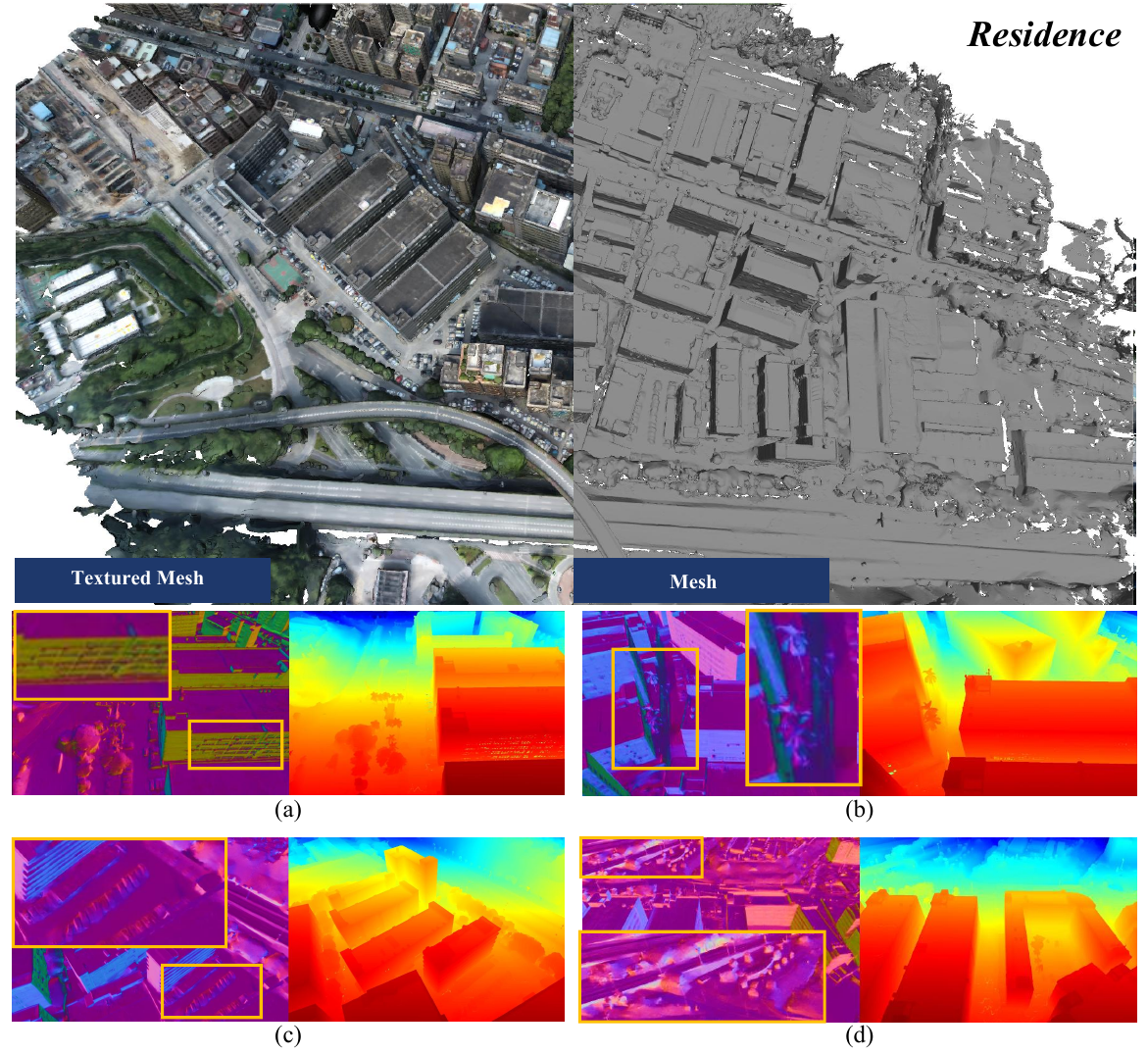}
    \caption{Qualitative analysis of the results of the Residence surface reconstruction.}
    \label{fig:residence_mesh}
\end{figure*}

\begin{figure}[t]
    \centering
    \includegraphics[width=1.0\linewidth]{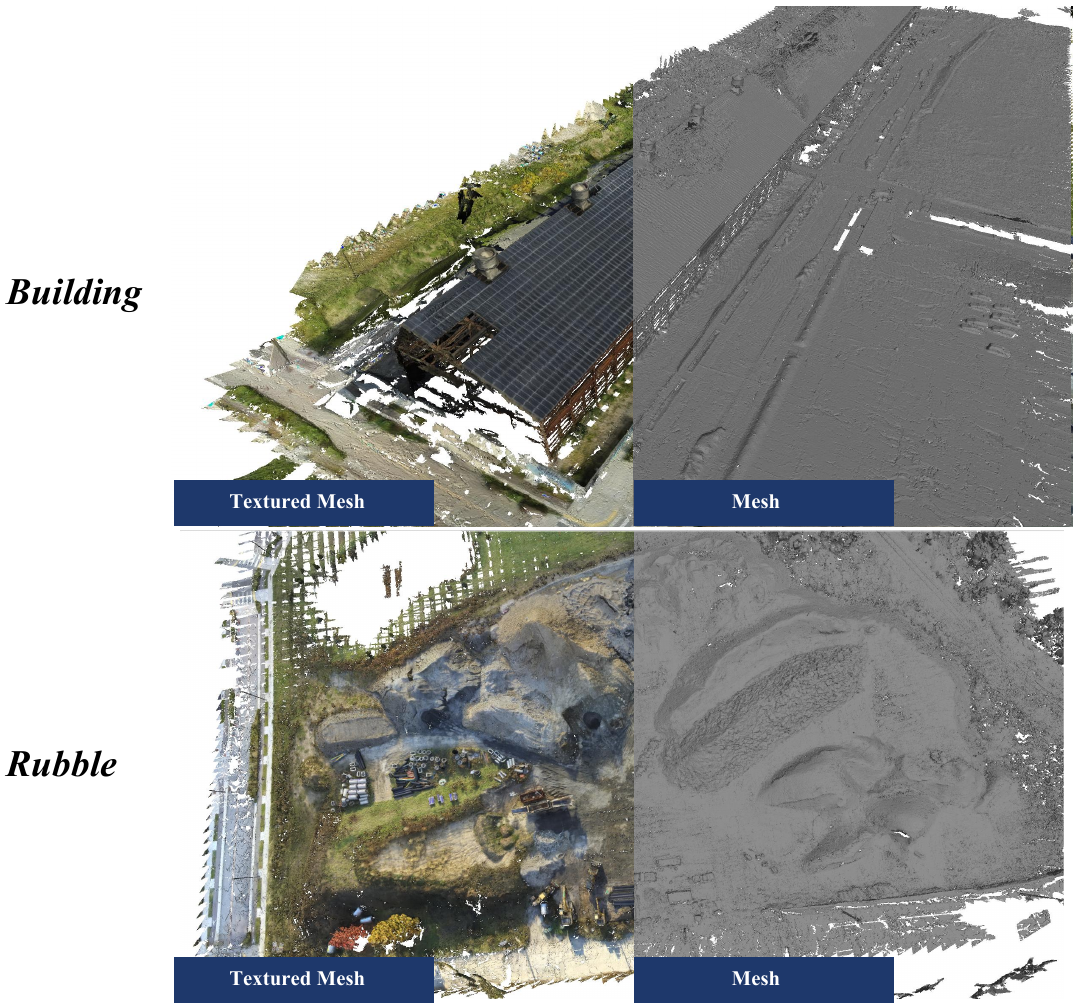}
    \caption{Qualitative analysis of the results of the Building and Rubble surface reconstruction.}
    \label{fig:building_rubble}
\end{figure}

\textbf{Results on Mill-19 and UrbanScene3D:}
Due to the lack of geometric ground truth and the inability of methods like 2DGS and GOF to complete training on these extremely large scenes, we lack direct comparative methods. Therefore, we conduct a qualitative analysis of the surface reconstruction results. As shown in Fig. \ref{fig:residence_mesh}, our reconstructed scenes are generally complete with smooth surfaces. On a more detailed level, our method effectively captures fine details. For instance, in the normal map shown in Fig. \ref{fig:residence_mesh}(a), not only are the buildings reconstructed, but finer details such as windows and air conditioning units are also well represented. In Fig. \ref{fig:residence_mesh}(b), the vegetation on the ground is also successfully reconstructed. Fig. \ref{fig:residence_mesh}(c) demonstrates that even small objects on the ground, such as vehicles, can be reconstructed. The most remarkable result is seen in Fig. \ref{fig:residence_mesh}(d), where the thin streetlights along the road are clearly reconstructed, highlighting the excellent performance of our method in capturing fine details.
We also apply our method to the Building and Rubble scenes, as shown in Fig. \ref{fig:building_rubble}, the results are equally impressive.

\subsubsection{Novel View Synthesis}\quad

Following the work of \cite{zhenxing2022switch} \cite{liu2024citygaussian} \cite{xu2023grid}, we use the peak signal-to-noise ratio (PSNR), structural similarity index measure (SSIM), and the learned perceptual image patch similarity (LPIPS) metrics to evaluate the quality of the rendered image.

\textbf{Results on WHU-OMVS:} We assess the rendering quality of the proposed method on the WHU-OMVS\cite{liu2023deep} dataset, as shown in Table \ref{tb:OMVS_render}. The results show that the proposed method significantly outperforms others in the LPIPS metric, with approximately a 46\% improvement over 3DGS. This indicates that the proposed method achieves superior visual consistency with the ground truth images. While the PSNR score is slightly lower than that of 3DGS—primarily due to the use of lighting compensation \cite{lin2024vastgaussian} in our method, which affects image brightness—our score remains competitive. Additionally, our SSIM score is also very competitive, only 0.002 lower than the best-performing method. 
\begin{table}
    \centering
    \caption{The image quality of novel view synthesis (NVS) on the WHU-OMVS dataset was compromised. The best results are highlighted in \textbf{bold}, and the second-best results are underlined.}
    \begin{tabular}{>{\centering\arraybackslash}p{2.5cm} | >{\centering\arraybackslash}p{1.5cm} >{\centering\arraybackslash}p{1.5cm} >{\centering\arraybackslash}p{1.5cm}}
        \hline
        method & SSIM↑ & PSNR↑ & LPIPS↓ \\ 
        \hline
         GOF & 0.925 & 29.07 & 0.666 \\ 
         3DGS & \textbf{0.940} & \textbf{31.14} & \underline{0.514} \\  
         2DGS & 0.927 & 29.89 & 0.668 \\
         proposed & \underline{0.938} & \underline{30.46} & \textbf{0.270} \\
        \hline
    \end{tabular}
    \label{tb:OMVS_render}
\end{table}

\textbf{Results on Mill-19 and UrbanScene3D:} Following the work of \cite{zhenxing2022switch} \cite{turki2022mega} and \cite{liu2024citygaussian}, we conduct experiments on the Mill-19 and UrbanScene3D datasets to further validate the rendering quality. As shown in Table \ref{tb:UrbanScene3D_render}, the proposed method achieves the highest scores across all three metrics (PSNR, SSIM, and LPIPS) in the Building and Rubble datasets, with PSNR significantly surpassing other methods. This demonstrates the method's ability to deliver optimal perceptual quality and achieve high-fidelity rendering. In the Residence dataset, the proposed method achieves the best results in both PSNR and LPIPS. Although our SSIM score is slightly lower than that of CityGaussian, it still significantly outperforms other methods, confirming the method's effectiveness across different datasets.

\begin{table*}
    \centering
    \caption{Quantitative comparison of rendered image on three datasets. The best results are highlighted in \textbf{bold}, and the second-best results are underlined.}
    \begin{tabular}{>{\centering\arraybackslash}p{2cm} | >{\centering\arraybackslash}p{1cm} >{\centering\arraybackslash}p{1cm} >{\centering\arraybackslash}p{1cm} | >{\centering\arraybackslash}p{1cm} >{\centering\arraybackslash}p{1cm} >{\centering\arraybackslash}p{1cm} | >{\centering\arraybackslash}p{1cm} >{\centering\arraybackslash}p{1cm} >{\centering\arraybackslash}p{1cm}}
        \hline
        \quad& \quad & Residence & \quad & \quad & Building & \quad & \quad & Rubble &\quad\\
        \hline
        method & SSIM↑ & PSNR↑ & LPIPS↓ & SSIM↑ & PSNR↑ & LPIPS↓ & SSIM↑ & PSNR↑ & LPIPS↓\\ 
        \hline
        MegaNeRF	&0.628&22.08&0.489&0.569&21.48&0.378&0.553&24.06&0.516\\
        Switch-NeRF	&0.654&\underline{22.57}&0.457&0.594&\underline{22.07}&0.332&0.562&24.31&0.496\\
            GP-NeRF	&0.661&22.31&0.448&0.566&21.03&0.486&0.565&24.06&0.496\\
     CityGaussian	&\textbf{0.813}&22.00&\underline{0.211}&\underline{0.778}&21.55&\underline{0.246}&\underline{0.813}&\underline{25.77}&\underline{0.228}\\
         proposed	&\underline{0.756}&\textbf{22.63}&\textbf{0.182}&\textbf{0.803}&\textbf{24.31}&\textbf{0.148}&\textbf{0.827}&\textbf{27.32}&\textbf{0.143}\\ 
        \hline
    \end{tabular}
    \label{tb:UrbanScene3D_render}
\end{table*}
\subsection{Ablation}

We conduct ablation experiments mainly on the WHU-OMVS dataset.

\begin{figure}[t]
    \centering
    \includegraphics[width=1\linewidth]{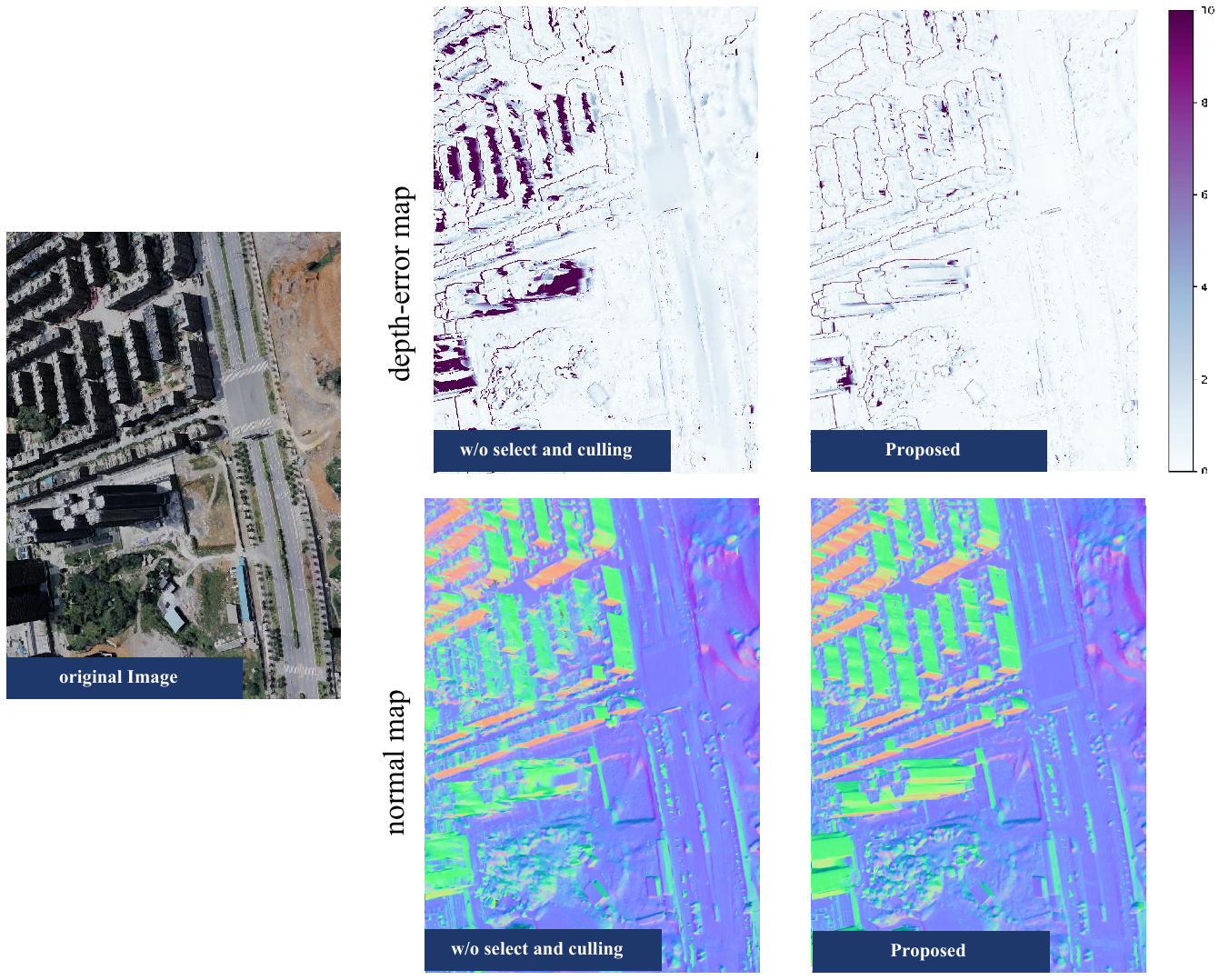}
    \caption{Qualitative comparison with and without viewpoint selection and culling.}
    \label{fig:select_ablation}
\end{figure}

\textbf{Viewpoint Selection and Culling:} As shown in Table \ref{tb:sel_abl}, disabling the viewpoint selection and culling strategy results in a significant reduction in reconstruction accuracy for the proposed method. This decline primarily stems from the insufficient number of views. As depicted in Fig. \ref{fig:select_ablation}, the lack of proper viewpoints leads to severe errors in localized areas of the depth error map, and certain parts of buildings in the normal map appear transparent and under-optimized. Thus, viewpoint selection and culling are crucial components of the chunking strategy, ensuring better coverage and optimization.
\begin{table*}
    \centering
    \caption{Ablation study of viewpoint selection and culling.}
    \begin{tabular}{>{\centering\arraybackslash}p{3cm} | >{\centering\arraybackslash}p{2cm} >{\centering\arraybackslash}p{2cm} >{\centering\arraybackslash}p{2cm} >{\centering\arraybackslash}p{2cm} >{\centering\arraybackslash}p{2cm}}
        \hline
        method & PAG$_{0.6m}$(\%) & PAG$_{0.8m}$(\%) & PAG$_{1.0m}$(\%) & MAE(m) & RMSE(m)\\ 
        \hline
         w/o VSC &69.16&	75.69	&79.65&	0.701&	1.449\\
         proposed & \textbf{82.58} & \textbf{87.07} & \textbf{89.50} & \textbf{0.451} & \textbf{1.012}\\
        \hline
    \end{tabular}
    \label{tb:sel_abl}
\end{table*}

\textbf{Ray-Gaussian Interaction:} As shown in Table \ref{tb:rgi}, the introduction of ray-gaussian interaction significantly improves reconstruction quality, evidenced by a 19.60\% improvement in $PAG_{0.6m}$. In the original 3DGS method \cite{kerbl20233d}, depth maps are generated using the initialized depth from Gaussians, and normal vector information is unavailable. In contrast, our approach accurately captures both depth and normal vectors, enabling the application of geometric constraints. As a result,the ray-gaussian intersection effectively mitigates the challenges associated with surface reconstruction, particularly those caused by the irregular distribution of Gaussian primitives.

\begin{table*}
    \centering
    \caption{Ablation experiments on the WHU-OMVS dataset. }
    \begin{tabular}{>{\centering\arraybackslash}p{3cm} | >{\centering\arraybackslash}p{2cm} >{\centering\arraybackslash}p{2cm} >{\centering\arraybackslash}p{2cm} >{\centering\arraybackslash}p{2cm} >{\centering\arraybackslash}p{2cm}}
        \hline
        method & PAG$_{0.6m}$(\%) & PAG$_{0.8m}$(\%) & PAG$_{1.0m}$(\%) & MAE(m) & RMSE(m)\\ 
        \hline
         w/o RGI &62.98	&71.72	&77.60	&0.757	&1.328\\
         w/o MVGC &81.74	&86.07	&88.40	&0.464	&1.068\\
         proposed & \textbf{82.58} & \textbf{87.07} & \textbf{89.50} & \textbf{0.451} & \textbf{1.012}\\
        \hline
    \end{tabular}
    \label{tb:rgi}
\end{table*}

\textbf{Multi-View Geometric Consistency Constraints:} The original 3DGS \cite{kerbl20233d} applies loss within a single view, which can lead to overfitting and fails to ensure consistency across multiple viewpoints. To address this issue, we introduce multi-view geometric consistency constraints. Table \ref{tb:rgi} demonstrates that the multi-view geometric consistency constraints effectively enhance reconstruction accuracy by ensuring coherence across multiple views.

\section{Discussion}
\textbf{Application Prospect:} The proposed method achieves geometric accuracy comparable to conventional open-source methods like Colmap and OpenMVS. However, it should be noted that we currently can only render a 1080p image (and depth map). In \cite{liu2024citygaussian}\cite{turki2022mega} and this work, the images are downsampled by a factor of four. This presents a barrier to applying 3DGS-based methods to images with full resolution. Future work must explore efficient methods for rendering higher-resolution images. Nevertheless, a significant advantage of the 3DGS-based method over traditional MVS approaches lies in its ability to render high-fidelity images while reconstructing surfaces. This offers new potential for surveying applications, allowing for measurements not only from the reconstructed mesh but also from the high-fidelity rendered images. Further research is required to develop a suitable measurement and evaluation method for these rendered images.

\textbf{Gaussian Seeds:} The 3DGS-based methods rely heavily on sparse point clouds generated through SfM as the initial seeds for Gaussian primitives. Datasets like Tianjin, which contain extensive textureless regions, present significant challenges due to the absence of initialized Gaussian primitives in these areas. The densification process attempts to densify Gaussian primitives into the textureless regions. However, these primitives, guided solely by RGB images, do not accurately reflect the actual surface. Due to the use of $\alpha$-blending for image rendering, these primitives in weak texture or textureless regions will interfere with the depth estimation of surrounding areas, leading to more depth estimation errors. As a result, the proposed method performs suboptimally on the Tianjin dataset. There remains considerable room for improvement in our approach. Future work could focus on improving SfM techniques or developing specialized densification strategies tailored for weak texture or textureless regions in 3DGS-based methods.
\section{Conclusion}
In this paper, we present the AGS framework, the first framework to achieve large-scale high-precision surface reconstruction from aerial images using a 3DGS-based approach. The proposed method combines a data chunking strategy specifically designed for aerial images, allowing each data block to be independently trained on a GPU and merged after training. Additionally, it incorporates the ray-gaussian intersection method to impose depth normal consistency constraints and multi-view geometric consistency constraints. We validate the geometric accuracy of our approach on the WHU-OMVS and Tianjin datasets and evaluate the rendering quality on the WHU-OMVS, Mill-19, and UrbanScene3D datasets. Experimental results demonstrate that the proposed method effectively performs surface reconstruction in large-scale scenes while achieving excellent rendering quality.

\bibliographystyle{unsrt}
\bibliography{ref.bib}

\end{document}